\documentclass[10pt, conference, letterpaper]{IEEEtran}
\IEEEoverridecommandlockouts

\usepackage{setspace}

\usepackage{booktabs} 
\usepackage{cite}
\usepackage{amsmath}
\usepackage{amssymb}
\usepackage{amsfonts}
\usepackage{float}
\usepackage{graphicx}
\usepackage{xcolor}
\usepackage{colortbl}
\usepackage{multirow}
\usepackage{mathtools}
\usepackage{algorithm}
\usepackage{algorithmicx}
\usepackage{algpseudocode}
\usepackage{tabto}
\usepackage{todonotes}
\usepackage{hhline}
\usepackage{enumitem}
\usepackage[belowskip=-8pt,aboveskip=8pt,labelfont={it,bf},font=small]{caption}
\def\BibTeX{{\rm B\kern-.05em{\sc i\kern-.025em b}\kern-.08em
    T\kern-.1667em\lower.7ex\hbox{E}\kern-.125emX}}

\usepackage{fancyhdr,lipsum}
\fancypagestyle{firstpage}{
  \fancyhf{}
  \fancyhead[L]{Published as a conference paper at IOLTS 2019}
}
\begin{document}
\title{TrISec: Training Data-Unaware Imperceptible Security Attacks on Deep Neural Networks}

 \author{
 \IEEEauthorblockN{Faiq Khalid$^1$, 
       Muhammad Abdullah Hanif$^1$,
 		Semeen Rehman$^1$,
 		Rehan Ahmed$^2$,
 		Muhammad Shafique$^1$}
  	\IEEEauthorblockA{$^1$\textit{Technische Universit\"at Wien (TU Wien), Vienna, Austria}\\
  	$^2$\textit{National University of Sciences and Technology (NUST), Islamabad, Pakistan}\\
  	Email: \{faiq.khalid,muhammad.hanif,semeen.rehman,muhammad.shafique\}@tuwien.ac.at, rehan.ahmed@seecs.edu.pk}
 }
\maketitle
\thispagestyle{firstpage}
\begin{abstract}
	Most of the data manipulation attacks on deep neural networks (DNNs) during the training stage introduce a perceptible noise that can be catered by preprocessing during inference, or can be identified during the validation phase. Therefore, data poisoning attacks during inference (e.g., adversarial attacks) are becoming more popular. However, many of them do not consider the imperceptibility factor in their optimization algorithms, and can be detected by correlation and structural similarity analysis, or noticeable (e.g., by humans) in multi-level security system. Moreover, majority of the inference attack rely on some knowledge about the training dataset. In this paper, we propose a novel methodology which automatically generates imperceptible attack images by using the back-propagation algorithm on pre-trained DNNs, without requiring any information about the training dataset (i.e., completely training data-unaware). We present a case study on traffic sign detection using the VGGNet trained on the German Traffic Sign Recognition Benchmarks dataset in an autonomous driving use case. Our results demonstrate that the generated attack images successfully perform misclassification while remaining imperceptible in both ``subjective'' and ``objective'' quality tests. \\ 
\end{abstract}

\section{Introduction}\label{introduction}
\begin{figure}[!t]
	\centering
	\includegraphics[width=1\linewidth]{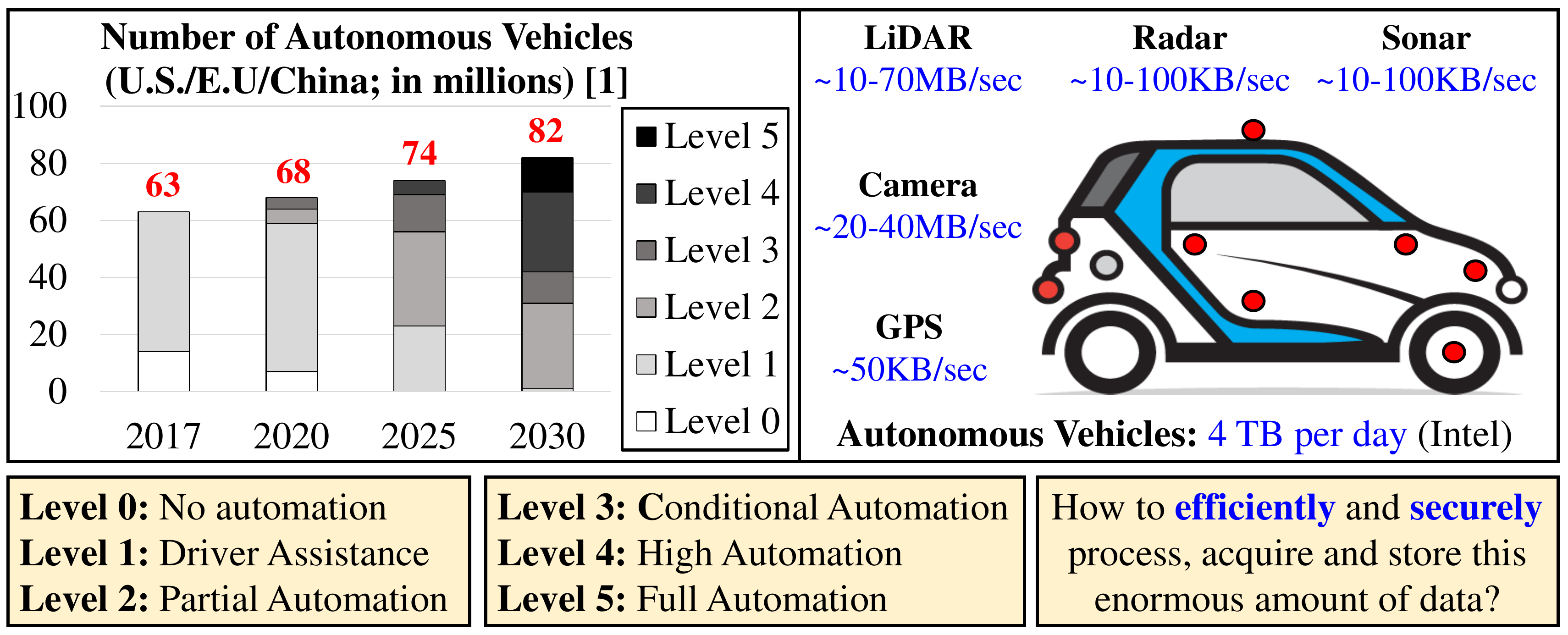} 
	\caption{\textit{Increasing trend of automation in self-driving cars and the expected amount of data generated per day in autonomous vehicles (data source: \cite{bansal2017forecasting}).}}
	\vskip -0.05in
	\label{fig:ML_Trend}
\end{figure}

The rapid development in technologies for smart cyber-physical systems is playing a vital role in the emergence of autonomous vehicles. For example, the number of autonomous vehicles in US, China and Europe is increasing exponentially, as shown in Fig. \ref{fig:ML_Trend}, and is estimated to reach approximately 80 million by 2030  \cite{bansal2017forecasting,lund2014worldwide,ratasich2019roadmap}. The amount of data generated by the multiple sensor nodes, e.g., LiDAR, navigation, camera, radar, etc., is massive (4 terabytes per day, see Fig. \ref{fig:ML_Trend}). To efficiently handle this data, the following research challenges need to be addressed:
 \begin{enumerate}[leftmargin=*]
     \item 	How to efficiently process the large amount of generated data with minimum energy consumption? 
     \item How to increase the storing capability to store this large amount of data in interpretable form while meeting the defined energy constraints?
 \end{enumerate}
Hence, there is a dire need to develop the computing architectures, methodologies, frameworks, algorithms and tools for processing the data in autonomous vehicles. Machine learning (ML) algorithms, especially deep neural networks (DNNs), serve as a prime solution because of their ability to effectively process the big data to solve tough problems in recognition, mining and synthesis applications~\cite{shafique2018overview}. DNNs in autonomous vehicles not only address the huge data mining challenges but they have also revolutionized several aspects of autonomous vehicles \cite{bojarski2016end}, e.g., obstacle detection, traffic sign detection, etc.

\subsection{Security Threats in DNN Modules}
\begin{figure}[!t]
	\centering
	\includegraphics[width=1\linewidth]{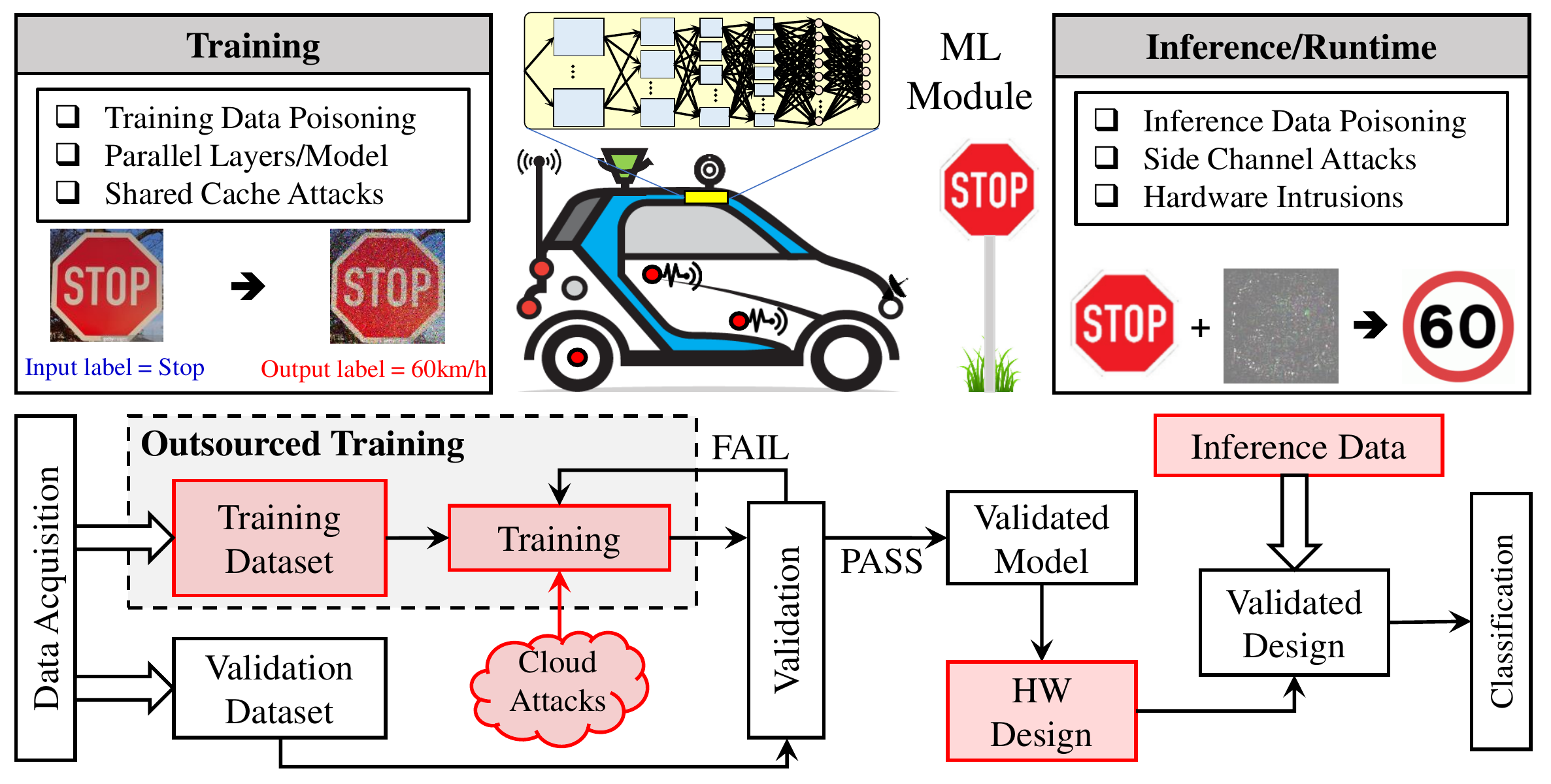}
	\caption{\textit{An overview of security threats and their respective payloads for ML algorithms during training and inference.}}
	\vskip -0.1in
	\label{fig:ML_Sec}
\end{figure}

\begin{figure*}[!t]
	\centering
	\includegraphics[width=1\linewidth]{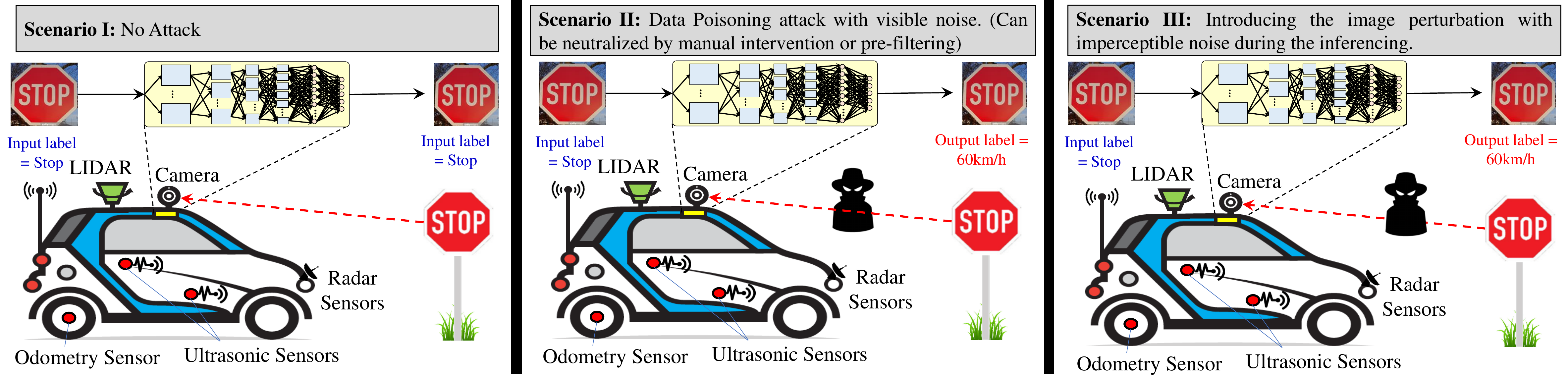} 
	\caption{\textit{Attack scenarios and corresponding threat models. In both scenarios, an attacker is adding an attack noise in the inference data. However, the noise is visible in scenario II but in scenario III the attack noise is not visible.}}
	\vskip -0.15in
	\label{fig:scenarios}
\end{figure*}

Several critical aspects, i.e., collision avoidance, traffic sign detection, and navigation with path following, are based on ML \cite{bojarski2016end}. These aspects are vulnerable to several security threats, as shown in Fig. \ref{fig:ML_Sec}, due to the unpredictability of the computations in the hidden layers of these DNN-based algorithms \cite{goodfellow2014generative}. As a result, autonomous vehicles are becoming more vulnerable to several security threats \cite{papernot2016limitations}. For instance, misclassification in object detection or traffic sign detection may lead to catastrophic incidents \cite{steinhardt2017certified}\cite{favaro2017examining}. Fig. \ref{fig:scenarios} shows two scenarios where an attacker can execute traffic sign misclassification. 

Unlike traditional systems, development of DNN-based systems comprise of three stages, i.e., training, hardware implementation, and inference in real-time. 
Each stage possesses its own security vulnerabilities \cite{papernot2016limitations}, such as data manipulation and corresponding payloads (e.g., confidence reduction, defined as an ambiguity in classification, and random or targeted misclassification) \cite{papernot2018sok}\cite{shafique2018overview}, as shown in Fig.~\ref{fig:ML_Sec}. 
During the training stage \cite{zhao2018data}\cite{wang2018data}, dataset \cite{shafahi2018poison}, tools and architecture/model are vulnerable to security attacks, such as adding parallel layers or neurons \cite{zou2018potrojan}, to perform security attacks \cite{rosenberg2017generic}. 
Similarly, during the hardware implementation and inference stages, computational hardware and real-time data can be exploited to perform security attacks \cite{kurakin2016adversarial}\cite{khalid2018security}.

{\em In the context of autonomous vehicles, data poisoning is one of the most commonly used attack on ML-modules.}  Typically, these attacks can be performed in two different ways:
\begin{enumerate}[leftmargin=*]
    \item {\bf Training Data Poisoning (TDP):} This attack introduces small patterned noise in training data samples to train the network for that particular noise pattern\cite{jagielski2018manipulating}. 
    However, for successful execution, the attacker requires complete access to the training dataset. 
    \textit{In most of these techniques, the noise pattern is perceptible and can be nullified by correlation and structural similarity analysis, or noticeable (e.g., by humans) in multi-level security system.} 
    
    \item {\bf Inference Data Poisoning (IDP):} This attack exploits the black-box/white-box model of the ML-modules to generate a noise patterns which can result in misclassification or confidence reduction \cite{papernot2017practical}\cite{papernot2016cleverhans,khalid2018fademl,khalid2018fademl2,khalid2019red}. 
    However, these noise patterns may \cite{gu2017badnets} or may not be imperceptible. 
    Several well-known imperceptible IDP attacks are limited-memory Broyden-Fletcher-Goldfarb-Shanno (L-BFGS) method \cite{szegedy2013intriguing}, Fast Gradient Sign Method (FGSM) \cite{rozsa2016adversarial} method, Jacobian-based Saliency Map Attack (JSMA) \cite{papernot2016limitations}, etc. Although \textit{adversarial IDP attacks generate imperceptible adversarial examples, they posses the following key limitations:} 
    \begin{enumerate}
        \item Most of them require reference sample(s) from the dataset.
        \item The imperceptibility of these attacks is achieved by \textit{only a single scaling factor} which is multiplied with the noise. This increases the probability of these attacks being detected by correlation and structural similarity analysis.  
    \end{enumerate}
\end{enumerate}
These limitations raise the fundamental research question: \textit{``How to generate an imperceptible attack noise pattern which can perform targeted or untargeted misclassification while ensuring the correlation coefficient and structural similarity at the specified maximal limit?''}

\subsection{Our Novel Contribution}
To address the above-mentioned research challenges, we propose an iterative methodology, \textit{TrISec}, to develop imperceptible adversarial examples, which does not require any knowledge of the training dataset and also incorporates the effects of attack noise on correlation coefficient and structural similarity index. \textbf{In a nutshell, this paper contributes the following:}
\begin{enumerate}[leftmargin=*]

    \item  To ensure training dataset independence, we propose to leverage the back-propagation algorithm on pre-trained DNNs to compute the change required in the input to perform targeted misclassification. 
    
    \item To ensure imperceptibility, we propose a two-step methodology, which first ensures correlation and then ensures high structural similarity to avoid high intensity noise at a particular location. 
\end{enumerate}
To evaluate the effectiveness of the proposed \textit{TriSec} attack, we evaluate it on the VGGNet, trained for the German Traffic Sign Recognition Benchmarks (GTSRB) dataset \cite{stallkamp2011german}. Our experimental results show that while ensuring the value of correlation coefficient and structural similarity around 1 and 0.999, respectively, the \textit{TriSec} attack successfully misclassifies a stop sign with almost 95\% confidence. Moreover, we also evaluate the perceptibility of the state-of-the-art attacks and compare it with the proposed \textit{TriSec} attack.

\section{Data Security Attacks on ML Inferencing}\label{ML_training}
The inference stage of ML algorithms have security vulnerabilities that are manipulation of the data acquisition block, communication channels and side-channel analysis to manipulate the inference data \cite{vorobeychik2018adversarial}\cite{joseph_nelson_rubinstein_tygar_2018}. Remote cyber-attacks and side-channel attacks have high computational costs and, therefore, are less frequently used \cite{papernot2017practical}\cite{shokri2017membership}. To design and implement the IDP attacks, we need to consider the following challenges:
\begin{enumerate}[leftmargin=*]
    \item How to relax the assumption of having access to the inference data acquisition block?
    \item How to generate an imperceptible attack noise pattern?
\end{enumerate}
To address these research challenges, several imperceptible IDPs have been proposed. In this section, we discuss and analyze the prominent adversarial attacks, like L-BFGS and FSGM.

\subsection{L-BFGS Method} This method generates an adversarial attack on DDNs \cite{szegedy2013intriguing}\cite{tabacof2016exploring}. The basic principle of the L-BFGS method is to iteratively optimize the attack noise with respect to the sample image, as shown in Equation~\ref{Eq:eq1}.
\begin{equation}
\min ||noise||_2 \implies f(x+noise) \neq f(x)
\label{Eq:eq1}
\end{equation} 
Where, the minimization of the $noise$ corresponds to imperceptibility. To illustrate the effectiveness of this method, we demonstrate this attack on the VGGNet trained for the GTSRB in Fig.~\ref{fig:adver_attack}. This experimental analysis shows that by introducing adversarial noise to the image, the input is missclassified, i.e., from a {\em stop} sign to a {\em speed limit 60km/h} sign. Although the L-BFGS method generates imperceptible adversarial example, it utilizes a basic linear search algorithm to optimize the noise, which makes it {\em computationally expensive and slow}.

\subsection{Fast Gradient Sign Method (FSGM)}
\begin{figure}[!t]
	\centering
	\includegraphics[width=0.9\linewidth]{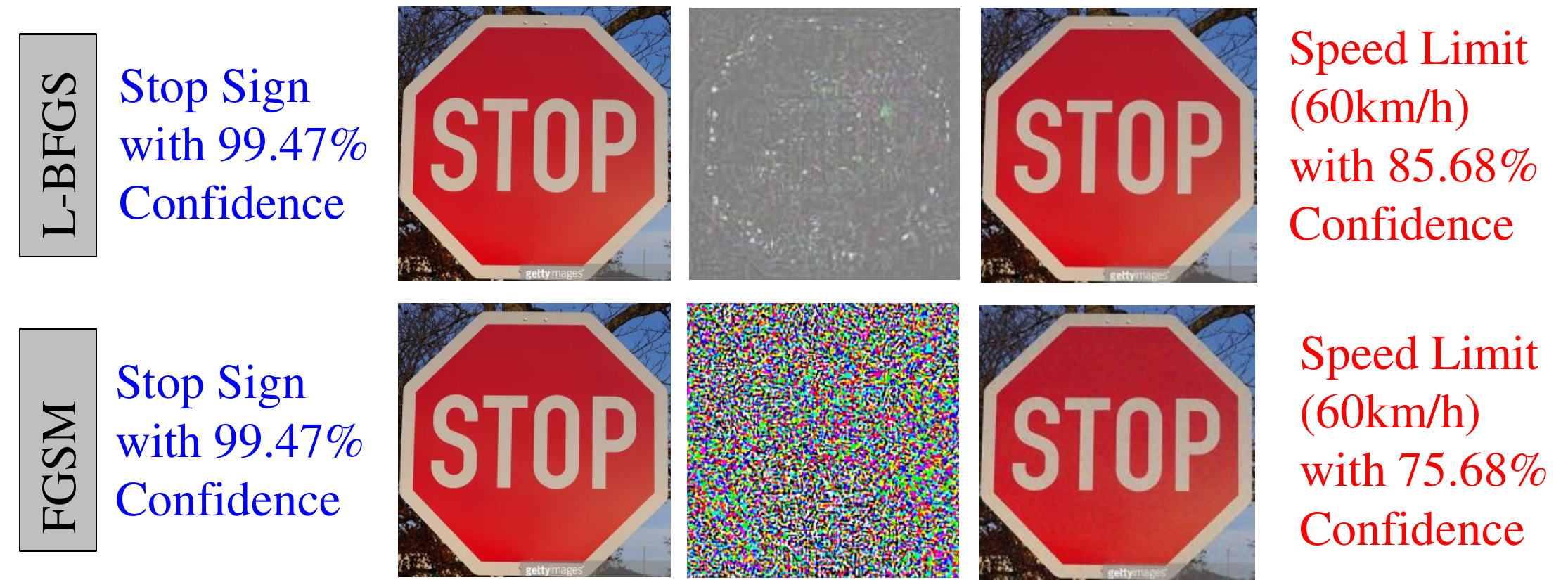} 
	\caption{\textit{Experimental Analysis for the L-BFGS and the FGSM attacks on the VGGNet trained for the GTSRB dataset.}}
	\vskip -0.1in
	\label{fig:adver_attack}
\end{figure}

To address the above-mentioned limitation of the L-BFGS, Goodfellow et al. proposed the FGSM attack to generate adversarial examples \cite{rozsa2016adversarial,goodfellow2015explaining}. The FSGM is faster and requires less computations because it performs one-step-gradient-update algorithm along the direction of the sign of the gradient at each pixel. 
Fig.~\ref{fig:adver_attack} shows an example of this attack on the VGGNet trained for the GTSRB dataset. The experimental analysis indicates that it can perform missclassifcation with imperceptible adversarial examples.

\subsection{Limitations}
Despite the above-mentioned attacks, there are many other adversarial attacks such as the basic iterative method (BIM) \cite{kurakin2018adversarial}, the JSMA \cite{papernot2016limitations}, the one-pixel attack \cite{su2017one}, the DeepFool attack \cite{moosavi2016deepfool}, the Zeroth Order Optimization (ZOO) attack \cite{chen2017zoo}, the CPPN EA Fool \cite{nguyen2015deep}, and the C\&W's attack \cite{carlini2017towards}. Though most of these attacks generate imperceptible noise patterns, they possess the following limitations:
\begin{enumerate}[leftmargin=*]
    \item Their optimization algorithms \textit{require reference sample(s)}; this limits their attack strength.
    \item Their optimization algorithms \textit{do not consider the perceptibility} (i.e., maximizing correlation coefficient and structural similarity index) in the optimization problem. 
\end{enumerate} 
These limitations raise the key research question: {\em How can we automatically generate an imperceptible inference data poisoning attack without having any knowledge about the reference sample(s).}

\section{TrISec: Training data-unaware imperceptible Attack methodology}\label{TriSec}
\begin{figure*}[!t]
	\centering
	\includegraphics[width=1\linewidth]{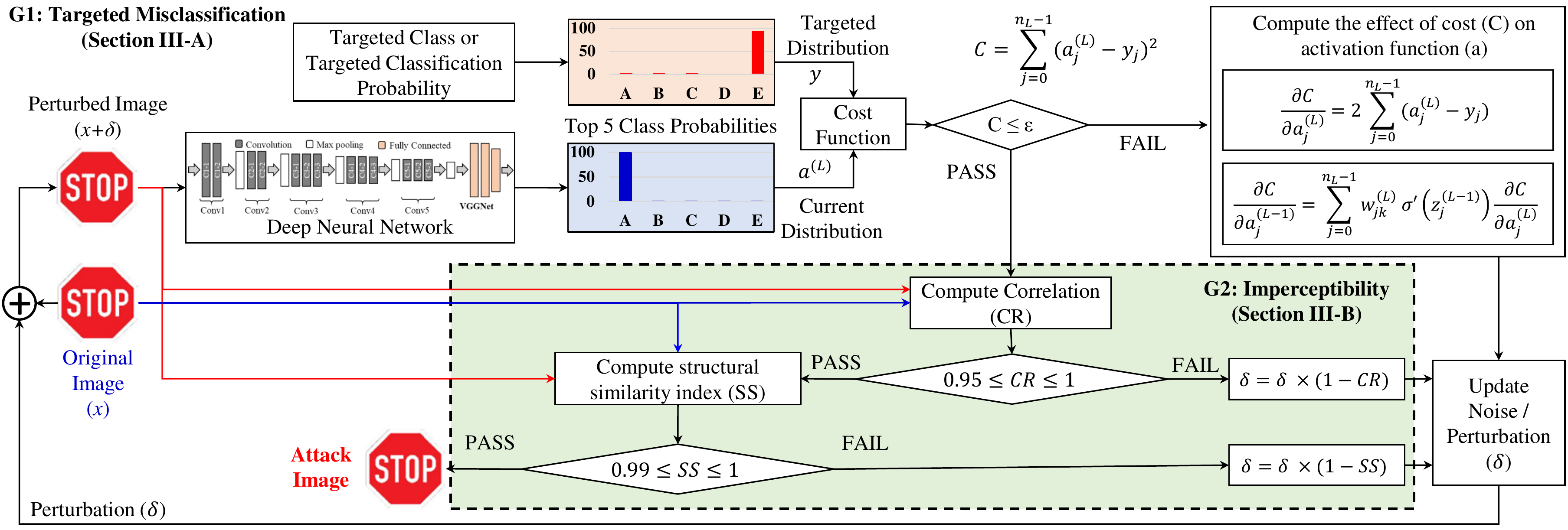} 
	\caption{\textit{Proposed methodology, Training data-unware Imperceptible Security Attack (\textit{TrISec}), which automatically generates imperceptible attack images for DNNs by leveraging the back-propagation algorithm.}}
	\vskip -0.15in
	\label{fig:ISA4ML}
\end{figure*}
To address the above-mentioned limitations (imperceptibility and dependency on training data samples), we propose a novel attack methodology, \textit{TrISec}, which leverages the back-propagation algorithm to optimize the perturbation in the attack image. \textbf{We formulate the following goals to achieve the imperceptible attack:} 
\begin{enumerate}[leftmargin=*]
    \item The first goal ($G1$) is to generate the noise/perturbation by computing and minimizing the cost function $C$ (Equation~\ref{Eq:eq4}) with respect to the probability of the targeted class, as formulated below:
    \begin{equation}
    \footnotesize
       G1 := \{min(C) \implies f(x+\delta) \neq f(x), f(x+\delta) = Y \}
        \label{Eq:eq2}
    \end{equation}
    Where, $f$, $x$ and $\delta$ represent the classification function, input image and generated noise/perturbation, respectively. 
    \item The second goal ($G2$) is to ensure the imperceptibility of the generated perturbation/noise in the input image, while maximizing the targeted class probability, as formulated below:
     \begin{equation}
     \footnotesize
     \begin{aligned}
       G2 := \{max(P(f(x+\delta)=Y)) \implies CR(x, x+\delta) \approx 1, \\ SSI(x, x+\delta) \approx 1\}
       \end{aligned}
        \label{Eq:eq3}
    \end{equation}
    Where, $CR(x, x+\delta)$, $SSI(x, x+\delta)$, $Y$ and $P(.)$ represent the correlation coefficient, the structural similarity index, targeted class and probability of a entity, respectively. 
\end{enumerate}
\subsection{Targeted Misclssification}
To achieve the first goal (G1), i.e., targeted misclassification, we propose the following steps, also shown in Fig.~\ref{fig:ISA4ML}.
\begin{enumerate}[leftmargin=*]
    
    \item We choose a classification probability distribution of the targeted class Y ($y = \{y_0, y_1, ...,y_{n_L-1}\}$, where $n_L$ represents the total number of classes, which is also the number of neurons in layer $L$), and compute the classification probability distribution of the perturbed image ($a^{(L)} = \{a^{(L)}_0, a^{(L)}_1, ...,a^{(L)}_{n_L-1}\}$). We compute the following cost function $C$ by using the difference between their respective classification probabilities: 
    \begin{equation}
      C = \sum_{j=0}^{n_L-1} (a_j^{(L)}-y_j)^2
    \label{Eq:eq4}
    \end{equation}
    
    \item To achieve the targeted misclassification with high confidence, we minimize the cost function by iteratively modifying the perturbation ($\delta$) in perturbed image. If the cost is greater than $\epsilon$ (a pre-defined small value), we back propagate this effect to the perturbed image by using the following set of equations.
    
    \textit{For back propagating the cost function from the last layer:}
    \begin{equation}
      \frac{\delta C}{\delta a^{(L)}}  = 2 \times \sum_{j=0}^{n_L-1}\Big(a^{(L)}_j-y_j\Big)
    \label{Eq:eq5}
    \end{equation}
    \textit{For rest of the layers (l):}
    \begin{equation}
      \frac{\delta C}{\delta a_{j}^{(l-1)}}  =  \sum_{j=0}^{n_l-1} \bigg(w_{jk}^{(l)} \times \sigma'(z_{j}^{(l-1)}) \times \frac{\delta C}{\delta a^{(L)}}\bigg) 
    \label{Eq:eq6}
    \end{equation}
    
    \begin{equation}
      z^{(l-1)}  =  \sum_{j=0}^{n_l-1} \bigg( w_{jk}^{(l-1)} \times a_{j}^{(l-1)} + b_{j} \bigg)
    \label{Eq:eq7}
    \end{equation}
    where, $a_{(l)}$, $w_{jk}^{(l)}$, $b$ and $\sigma$ are the output activations, the weights of layer $l$, biases and activation function, respectively. To understand the impact of the cost function on the activations, we compute the inverse effect of Equations~\ref{Eq:eq5} and~\ref{Eq:eq6}. We repeat this step until the cost function is less than or equal to $\epsilon$, to ensure the targeted misclassification. 
\end{enumerate}
\subsection{Imperceptibility}
To achieve the second goal (G2), i.e., to ensure the imperceptibility, we propose a two-step methodology (Fig~\ref{fig:ISA4ML}). 
\begin{enumerate}[leftmargin=*]
    \item After achieving the targeted misclassification, first, we compute the \textit{correlation coefficient (CR)} of the perturbed image and compare it with a pre-defined limit (i.e., in our case, $0.95 \leq CR \leq1 $). To maximize the \textit{CR}, we propose to modify the perturbation using the following equation. 
    \begin{equation}
        \delta = \delta \times (1-CR(x, x+\delta)) 
    \label{Eq:eq8}
    \end{equation}
    \item Although the correlation coefficient somewhat ensures the imperceptibility, it does not guarantee that the noise is scattered across the image (multiple pixels). Therefore, to ensure that the noise in not concentrated at a particular location in the image, we propose to compute the \textit{structural similarity index (SSI)} of the perturbed image and compare it with a pre-defined limit (i.e., in our case, $0.99 \leq SSI \leq1 $). To maximize the \textit{SSI} and imperceptibility, we propose to modify the perturbation using the following equation.
    \begin{equation}
        \delta = \delta \times (1-SSI(x, x+\delta)) 
    \label{Eq:eq9}
    \end{equation}
\end{enumerate}
By ensuring high \textit{CR} and \textit{SSI}, we ensure high imperceptibility of the attack noise. Note, the above pre-defined limits can be tightened as much as possible to satisfy imperceptibility requirements. However, it may result in more iterations of the methodology. Therefore, these limits also provide a tradeoff between the level of imperceptibility and the speed of the attack image generation. 
\section{Results and Discussions}
To illustrate the effectiveness of the proposed \textit{TrISec} attack, we evaluate it using the following experimental setup, and also compared it with few of the state-of-the-art adversarial attacks, i.e., the FGSM and the L-BFG, available in the open-source \textit{Cleverhans} library.
\begin{figure}[b]
	\centering
	\vskip -0.2in
	\includegraphics[width=1\linewidth]{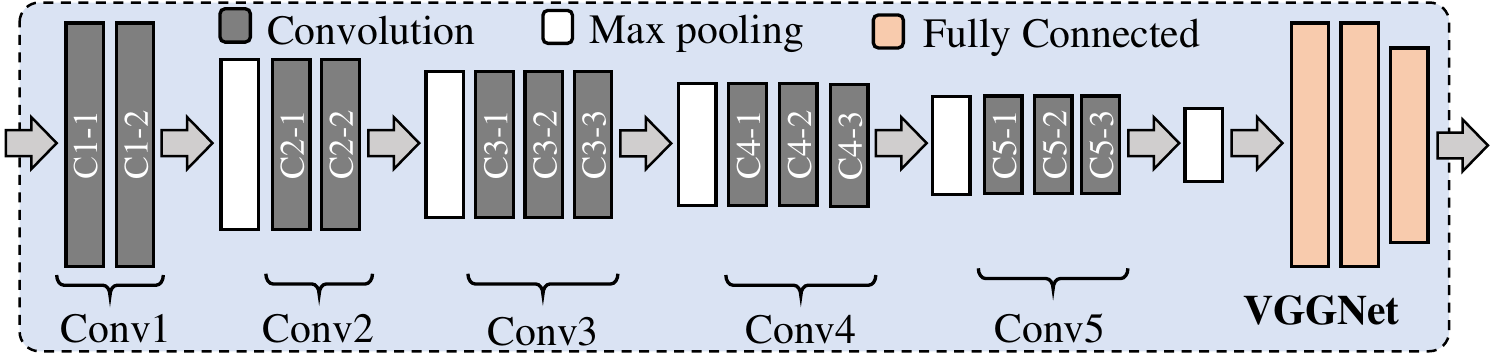}
	\caption{\textit{VGGNet \cite{khalid2018fademl}: Conv1 (64 filters), Conv2 (128 filters), Conv3 (256 filters), Conv4 (512 filters) and Conv5 (512 filters), where Convn is the $nth$ set of convolutional layers}}
	\label{fig:vggnet}
\end{figure}
\begin{figure*}[!t]
	\centering
	\includegraphics[width=1\linewidth]{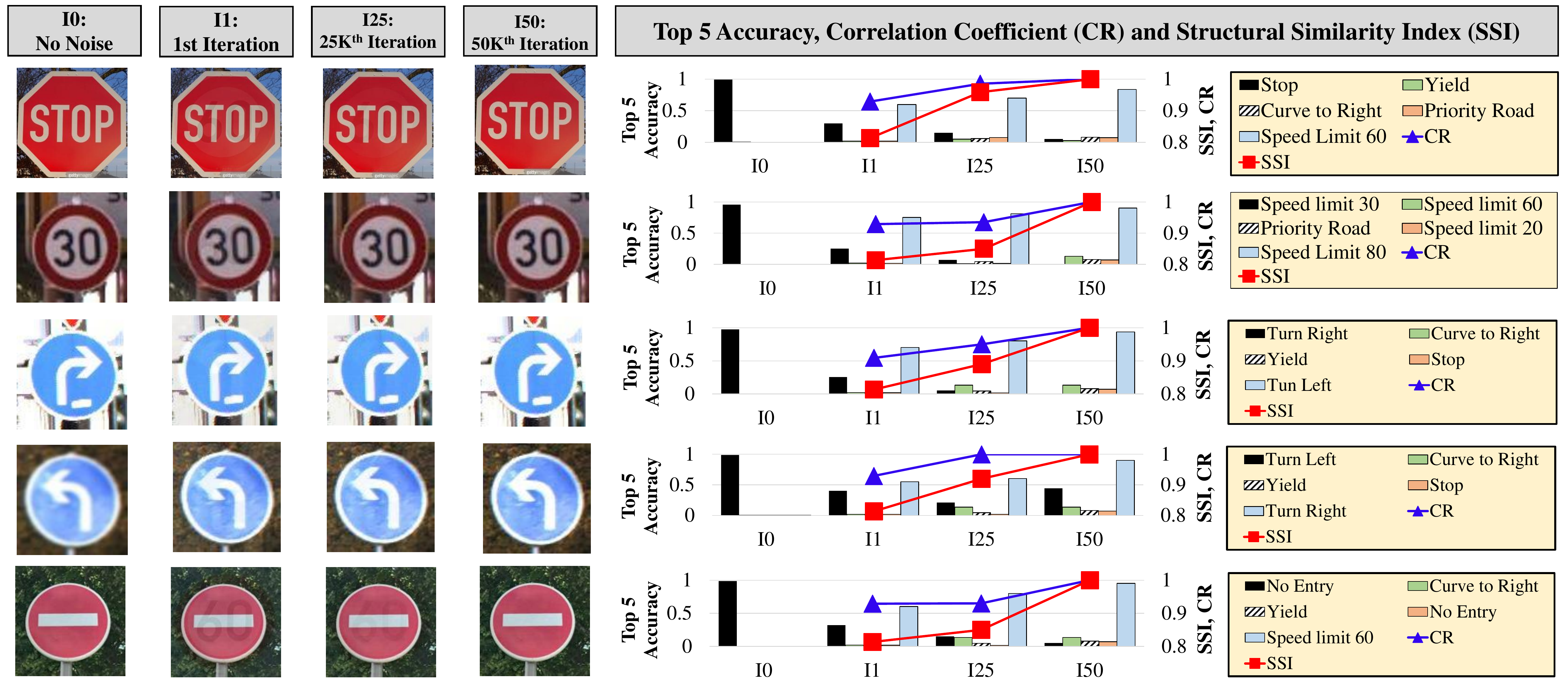} 
	\caption{\textit{Experimental results of \textit{TriSec} attack on the VGGNet, trained on the GTSRB dataset. This analysis shows that with an increase in iterations the attack noise become imperceptible while ensuring the targeted misclassification with high confidence.}}
	\label{fig:exp_ISA4ML}
\end{figure*}
\begin{figure*}[!t]
	\centering
	\includegraphics[width=1\linewidth]{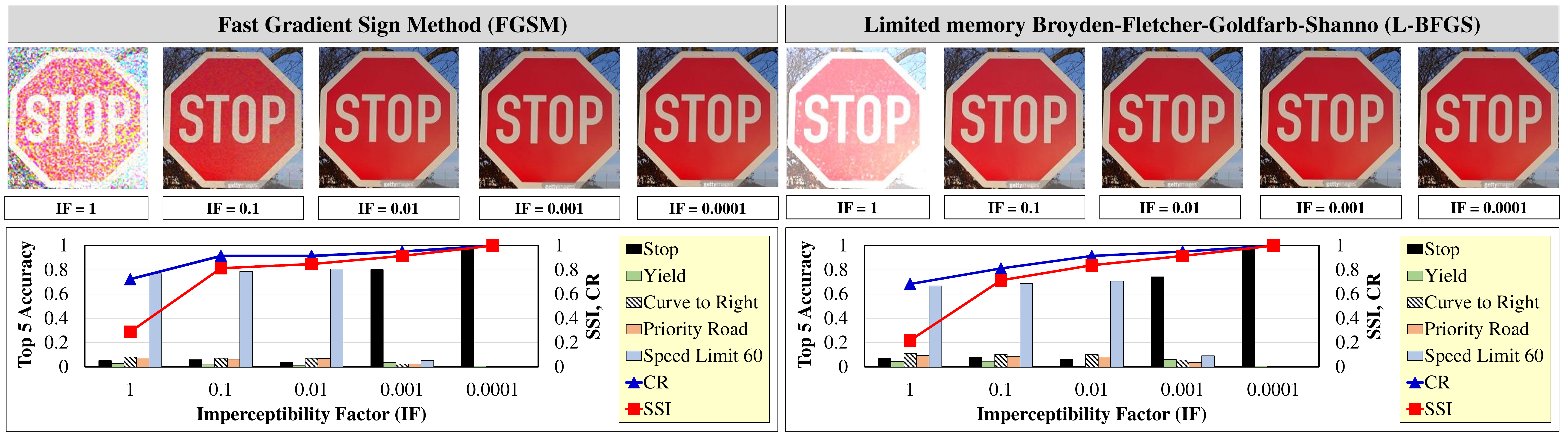} 
	\caption{\textit{Experimental analysis of the state-of-the-art adversarial attacks, i.e., the FGSM and the L-BFGS. This analysis shows the impact of the imperceptibility factor on the attack success. Imperceptibility factor (IF) is defined as the parameter to vary the impact of the attack noise on the correlation and structural similarity analysis. CR and SSI represent the correlation coefficient and structural similarity index, respectively.}}
	\vskip -0.1in
	\label{fig:comp}
\end{figure*}
\subsection{Experimental Setup used in Our Evaluation and Analysis}
\begin{enumerate}[leftmargin=*]
    \item \textbf{DNN and Dataset:} We use the VGGNet (Fig. \ref{fig:vggnet}) trained on the GTSRB dataset \cite{stallkamp2011german}.
    
   \item \textbf{Threat Model:} We assume one of the commonly used threat models (i.e., white-box), which states that an attacker has access to all the DNN parameters but cannot modify them. Also, the attacker does not have any access to the training dataset. 
   
   \item \textbf{\textit{TrISec} Settings:} We define the following constraints: 
   \begin{enumerate}
       \item Upper bound of cost as 0.05, i.e., $\epsilon = 0.0025$.
       \item Lower bound of CR as 0.95, i.e., $0.95 \leq CR \leq1 $.
       \item Lower bound of SSI as 0.99, i.e., $0.99 \leq SSI \leq1 $
   \end{enumerate}
\end{enumerate}

\subsection{Evaluation and Discussion}
The experimental results in Fig.~\ref{fig:exp_ISA4ML} show how the imperceptibility is achieved over iterations. In this analysis, we make the following key observations: 
\begin{enumerate}[leftmargin=*]
    \item After the first iteration, we achieve the targeted miscalssification but the intensity of the attack noise is very high and is clearly visible in the images of Fig.~\ref{fig:exp_ISA4ML} under ``I1'' label. Moreover, the corresponding values of $CR$ and $SSI$ are $0.901$ and $0.8133$ which are below the defined bounds, i.e., $0.95$ and $0.99$. 
    \item The probability distribution of the top-5 classes of the considered input images and the corresponding values of $CR$ and $SSI$ after different iterations of \textit{TrISec} methodology are presented in the analysis graphs of Fig.~\ref{fig:exp_ISA4ML}. These graphs show that with an increase in imperceptibility ($CR$ and $SSI$), the probability of the targeted class increases. 
\end{enumerate}

\subsection{Comparison with State-of-the-art Adversarial Attacks}
To demonstrate the effectiveness of \textit{TrISec}, in this section, we present analysis in the form of a trade-off between imperceptibility (i.e., CR and SSI) and attack success for the state-of-the-art adversarial attacks like the FGSM and the L-BFGS (available in the open-source \textit{Cleverhans} library), and \textit{TrISec}. To vary the imperceptibility of attack noise generated by the FGSM and the L-BGFS, we introduce a factor, \textit{imperceptibility factor} (IF), which is multiplied with the attack noise before adding to the input image to generate the adversarial example. For this analysis, we consider the VGGNet, trained on the GTSRB dataset. Fig. \ref{fig:comp} illustrates the effects of varying IF on attack success of the FGSM and the L-BFGS attacks. 

By analyzing the experimental results presented in Fig.~\ref{fig:comp}, we observe that in the FGSM, if the $IF$ is ``1'' (which represents attack noise being added without any scaling), the misclassifcation goal is achieved but the correlation coefficient and structural similarity index are ``0.72'' and ``0.29'', respectively. If we decrease the $IF$ the imperceptibility increases, however, the attack remains successful till $IF = 0.01$. For example, at IF=0.001 and 0.0001, the FGSM fails but the correlation coefficient and structural similarity index  are ``100\%'' and ``99.699\%'', respectively. Similar trend is observed in case of the L-BFGS attack.

On the other hand, the proposed \textit{TrISec} attack incorporates the imperceptibility (i.e., correlation coefficient and structural similarity index) in its optimization goal (see Equations \ref{Eq:eq2} and \ref{Eq:eq3}). Therefore, it can generate a successful attack with highly imperceptible noise without any imperceptibility factor, as shown in Fig. \ref{fig:exp_ISA4ML}.  

\subsection{Key Insight}
To ensure the maximum imperceptibility, ideally the attacker should only rely on one similarity metric but in practical scenarios this is not valid. For example, consider the example scenario (stop sign being mapped to speed limit 60km/h sign) in the first row of Fig.~\ref{fig:exp_ISA4ML}, at 25K iterations the $CR$ value of is greater then the defined bound (i.e., $0.95$), however, the noise is not imperceptible. \textit{Thus, it is required to consider multiple kinds of similarity metrics to ensure imperceptibility.} In short, in ML sub-systems with multi-level security, for instance, where a cross-correlation and structural similarity based checks are employed in a ML sub-system, our technique would succeed, while state-of-the-art attacks like the FGSM and the L-BFGS will fail.
    
\section{Conclusion}\label{conclusion}
Typically, inference data poisoning attacks consider the white-box scenario while assuming that attacker has access to the DNN parameters but cannot modify them. 
However, most of these attacks do not consider imperceptibility with respect to objective quality metrics like correlation coefficient and structural similarity index, which can be crucial in a multi-level secure ML sub-systems. 
In this paper, we proposed a novel {\em training data-unaware} methodology to automatically generate {\em imperceptible adversarial examples} by leveraging the back-propagation algorithm on pre-trained deep neural networks (DNNs), as well as incorporating objective quality metrics inside the optimization loop for a robust attack image generation. 
We successfully demonstrated the proposed \textit{TrISec}, on one of the state-of-the-art DNN, {\em VGGNet},  deployed in an autonomous driving use-case for traffic sign detection. 
Our experiments showed that the generated attacks go unnoticed in both subjective and objective tests, with close to ideal correlation and structural similarity index with respect to the clean input image. 
Our study showed that such attacks can be very powerful and would require new security-aware design methods for developing robust machine learning-based systems for autonomous vehicles. 
\section*{Acknowledgement}
This work was partially supported by the Erasmus+ International Credit Mobility (KA107). 
\begin{spacing}{0.92}
\footnotesize 
\bibliographystyle{IEEEtran}
\bibliography{TrISec.bbl}

\begin{thebibliography}{10}
\providecommand{\url}[1]{#1}
\csname url@samestyle\endcsname
\providecommand{\newblock}{\relax}
\providecommand{\bibinfo}[2]{#2}
\providecommand{\BIBentrySTDinterwordspacing}{\spaceskip=0pt\relax}
\providecommand{\BIBentryALTinterwordstretchfactor}{4}
\providecommand{\BIBentryALTinterwordspacing}{\spaceskip=\fontdimen2\font plus
\BIBentryALTinterwordstretchfactor\fontdimen3\font minus
  \fontdimen4\font\relax}
\providecommand{\BIBforeignlanguage}[2]{{%
\expandafter\ifx\csname l@#1\endcsname\relax
\typeout{** WARNING: IEEEtran.bst: No hyphenation pattern has been}%
\typeout{** loaded for the language `#1'. Using the pattern for}%
\typeout{** the default language instead.}%
\else
\language=\csname l@#1\endcsname
\fi
#2}}
\providecommand{\BIBdecl}{\relax}
\BIBdecl

\bibitem{bansal2017forecasting}
P.~Bansal~et al., ``Forecasting americans’ long-term adoption of connected
  and autonomous vehicle technologies,'' \emph{Transportation Research Part A:
  Policy and Practice}, vol.~95, pp. 49--63, 2017.

\bibitem{lund2014worldwide}
D.~Lund~et al., ``Worldwide and regional internet of things (iot) 2014--2020
  forecast: A virtuous circle of proven value and demand,'' \emph{International
  Data Corporation (IDC), Tech. Rep}, vol.~1, 2014.

\bibitem{ratasich2019roadmap}
D.~Ratasich~et al., ``A roadmap toward the resilient internet of things for
  cyber-physical systems,'' \emph{IEEE Access}, vol.~7, pp. 13\,260--13\,283,
  2019.

\bibitem{shafique2018overview}
M.~Shafique~et al., ``An overview of next-generation architectures for machine
  learning: Roadmap, opportunities and challenges in the {IoT} era,'' in
  \emph{\textbf{DATE}}.\hskip 1em plus 0.5em minus 0.4em\relax IEEE, 2018, pp.
  827--832.

\bibitem{bojarski2016end}
M.~Bojarski~et al., ``End to end learning for self-driving cars,''
  \emph{arXiv:1604.07316}, 2016.

\bibitem{goodfellow2014generative}
I.~Goodfellow~et al., ``Generative adversarial nets,'' in \emph{NIPS}, 2014,
  pp. 2672--2680.

\bibitem{papernot2016limitations}
N.~Papernot~et al., ``The limitations of deep learning in adversarial
  settings,'' in \emph{\textbf{EuroS\&P}}.\hskip 1em plus 0.5em minus
  0.4em\relax IEEE, 2016, pp. 372--387.

\bibitem{steinhardt2017certified}
J.~Steinhardt~et al., ``Certified defenses for data poisoning attacks,'' in
  \emph{NIPS}, 2017, pp. 3520--3532.

\bibitem{favaro2017examining}
F.~M. Favar{\`o}~et al., ``Examining accident reports involving autonomous
  vehicles in california,'' \emph{PLoS one}, vol.~12, no.~9, p. e0184952, 2017.

\bibitem{papernot2018sok}
N.~Papernot~et al., ``So{K}: Security and privacy in machine learning,'' in
  \emph{\textbf{EuroS\&P}}.\hskip 1em plus 0.5em minus 0.4em\relax IEEE, 2018,
  pp. 399--414.

\bibitem{zhao2018data}
M.~Zhao~et al., ``Data poisoning attacks on multi-task relationship learning,''
  in \emph{\textbf{AAAI}}, 2018, pp. 2628--2635.

\bibitem{wang2018data}
Y.~Wang~et al., ``Data poisoning attacks against online learning,''
  \emph{arXiv:1808.08994}, 2018.

\bibitem{shafahi2018poison}
A.~Shafahi~et al., ``Poison frogs! targeted clean-label poisoning attacks on
  neural networks,'' \emph{arXiv:1804.00792}, 2018.

\bibitem{zou2018potrojan}
M.~Zou~et al, ``Potrojan: powerful neural-level trojan designs in deep learning
  models,'' \emph{arXiv preprint arXiv:1802.03043}, 2018.

\bibitem{rosenberg2017generic}
I.~Rosenberg~et al., ``Generic black-box end-to-end attack against rnns and
  other api calls based malware classifiers,'' \emph{arXiv:1707.05970}, 2017.

\bibitem{kurakin2016adversarial}
A.~Kurakin~et al., ``Adversarial examples in the physical world,''
  \emph{arXiv:1607.02533}, 2016.

\bibitem{khalid2018security}
F.~Khalid~et al., ``Security for machine learning-based systems: Attacks and
  challenges during training and inference,'' \emph{arXiv preprint
  arXiv:1811.01463}, 2018.

\bibitem{jagielski2018manipulating}
M.~Jagielski~et al., ``Manipulating machine learning: Poisoning attacks and
  countermeasures for regression learning,'' \emph{arXiv:1804.00308}, 2018.

\bibitem{papernot2017practical}
N.~Papernot~et al., ``Practical black-box attacks against machine learning,''
  in \emph{\textbf{AsiaCCS}}.\hskip 1em plus 0.5em minus 0.4em\relax ACM, 2017,
  pp. 506--519.

\bibitem{papernot2016cleverhans}
------, ``{CleverHans} v2. 0.0: an adversarial machine learning library,''
  \emph{arXiv:1610.00768}, 2016.

\bibitem{khalid2018fademl}
F.~Khalid~et al., ``{FAdeML:} understanding the impact of pre-processing noise
  filtering on adversarial machine learning,'' \emph{arXiv:1811.01444}, 2018.

\bibitem{khalid2018fademl2}
------, ``{FAdeML:} understanding the impact of pre-processing noise filtering
  on adversarial machine learning,'' in \emph{\textbf{{DATE}}}.\hskip 1em plus
  0.5em minus 0.4em\relax IEEE, 2019.

\bibitem{khalid2019red}
F.~Khalid, H.~Ali, M.~A. Hanif, S.~Rehman, R.~Ahmed, and M.~Shafique,
  ``{RED-Attack}: Resource efficient decision based attack for machine
  learning,'' \emph{arXiv preprint arXiv:1901.10258}, 2019.

\bibitem{gu2017badnets}
T.~Gu~et al., ``Badnets: Identifying vulnerabilities in the machine learning
  model supply chain,'' \emph{arXiv:1708.06733}, 2017.

\bibitem{szegedy2013intriguing}
C.~Szegedy~et al., ``Intriguing properties of neural networks,''
  \emph{arXiv:1312.6199}, 2013.

\bibitem{rozsa2016adversarial}
A.~Rozsa~et al., ``Adversarial diversity and hard positive generation,'' in
  \emph{\textbf{CVPR Workshop}}.\hskip 1em plus 0.5em minus 0.4em\relax IEEE,
  2016, pp. 25--32.

\bibitem{stallkamp2011german}
J.~Stallkamp~et al., ``The {G}erman traffic sign recognition benchmark: a
  multi-class classification competition,'' in \emph{\textbf{IJCNN}}.\hskip 1em
  plus 0.5em minus 0.4em\relax IEEE, 2011, pp. 1453--1460.

\bibitem{vorobeychik2018adversarial}
Y.~Vorobeychik~et al., ``Adversarial machine learning,'' \emph{Synthesis
  Lectures on AI and ML}, vol.~12, no.~3, pp. 1--169, 2018.

\bibitem{joseph_nelson_rubinstein_tygar_2018}
A.~D.~J. et~al., \emph{Adversarial Machine Learning}.\hskip 1em plus 0.5em
  minus 0.4em\relax Cambridge University Press, 2018.

\bibitem{shokri2017membership}
R.~Shokri~et al., ``Membership inference attacks against machine learning
  models,'' in \emph{\textbf{S\&P}}.\hskip 1em plus 0.5em minus 0.4em\relax
  IEEE, 2017, pp. 3--18.

\bibitem{tabacof2016exploring}
P.~Tabacof~et al., ``Exploring the space of adversarial images,'' in
  \emph{\textbf{IJCNN}}.\hskip 1em plus 0.5em minus 0.4em\relax IEEE, 2016, pp.
  426--433.

\bibitem{goodfellow2015explaining}
I.~Goodfellow~et al., ``Explaining and harnessing adversarial examples,''
  \emph{stat}, vol. 1050, p.~20, 2015.

\bibitem{kurakin2018adversarial}
A.~Kurakin~et al., ``Adversarial examples in the physical world,'' in
  \emph{\textbf{AISS}}.\hskip 1em plus 0.5em minus 0.4em\relax Chapman and
  Hall/CRC, 2018, pp. 99--112.

\bibitem{su2017one}
J.~Su~et al., ``One pixel attack for fooling deep neural networks,''
  \emph{arXiv:1710.08864}, 2017.

\bibitem{moosavi2016deepfool}
S.~Moosavi-Dezfooli~et al., ``Deepfool: a simple and accurate method to fool
  deep neural networks,'' in \emph{\textbf{CVPR}}.\hskip 1em plus 0.5em minus
  0.4em\relax IEEE, 2016, pp. 2574--2582.

\bibitem{chen2017zoo}
P.~Chen~et al., ``Zoo: Zeroth order optimization based black-box attacks to
  deep neural networks without training substitute models,'' in
  \emph{\textbf{AISec}}.\hskip 1em plus 0.5em minus 0.4em\relax ACM, 2017, pp.
  15--26.

\bibitem{nguyen2015deep}
A.~Nguyen~et al., ``Deep neural networks are easily fooled: High confidence
  predictions for unrecognizable images,'' in \emph{\textbf{CVPR}}.\hskip 1em
  plus 0.5em minus 0.4em\relax IEEE, 2015, pp. 427--436.

\bibitem{carlini2017towards}
N.~Carlini~et al., ``Towards evaluating the robustness of neural networks,'' in
  \emph{\textbf{S\&P}}.\hskip 1em plus 0.5em minus 0.4em\relax IEEE, 2017, pp.
  39--57.

\end{thebibliography}
\end{spacing}

\end{document}